\documentclass{article}
\usepackage{spconf,amsmath,graphicx}
\usepackage{booktabs}
\usepackage{dingbat}

\title{EGOFALLS: A visual-audio dataset and benchmark for fall detection using egocentric cameras}
\name{Xueyi Wang}
\address{University of Groningen\\xueyi.wang@rug.nl}
\begin{document}
%
\maketitle
\begin{abstract}

Falls are significant and often fatal for vulnerable populations such as the elderly. 
Previous works have addressed the detection of falls by relying on data capture by a single sensor, images or accelerometers. In this work, we rely on multimodal descriptors extracted from videos captured by egocentric cameras.  
Our proposed method includes a late decision fusion layer that builds on top of the extracted descriptors.
Furthermore, we collect a new dataset on which we assess our proposed approach. We believe this is the first public dataset of its kind.
The dataset comprises 10,948 video samples by 14 subjects. 
We conducted ablation experiments to assess the performance of individual feature extractors, fusion of visual information, and fusion of both visual and audio information.  Moreover, we experimented with internal and external cross-validation. 
Our results demonstrate that the fusion of audio and visual information through late decision fusion improves detection performance, making it a promising tool for fall prevention and mitigation. 

\end{abstract}
\section{Introduction}
\label{sec:intro}
Falls represent a major source of morbidity and mortality among the elderly population, often leading to both physical injuries and psychological consequences \cite{yavuz2010smartphone}. According to recent statistics reported by Haagsma et al. \cite{haagsma2020falls}, approximately 14\% of the adult population in Western countries have experienced fall-related injuries. Alarmingly, the incidence of falls within this demographic has surged by 54\% since 1990 \cite{haagsma2020falls}. Such trends not only place a considerable burden on families caring for elderly individuals who have sustained injuries, but also have significant implications for the financial health of nations, as they strain healthcare systems and associated resources.

\begin{figure}[h]
    \centering
    \includegraphics[width=0.5\textwidth]{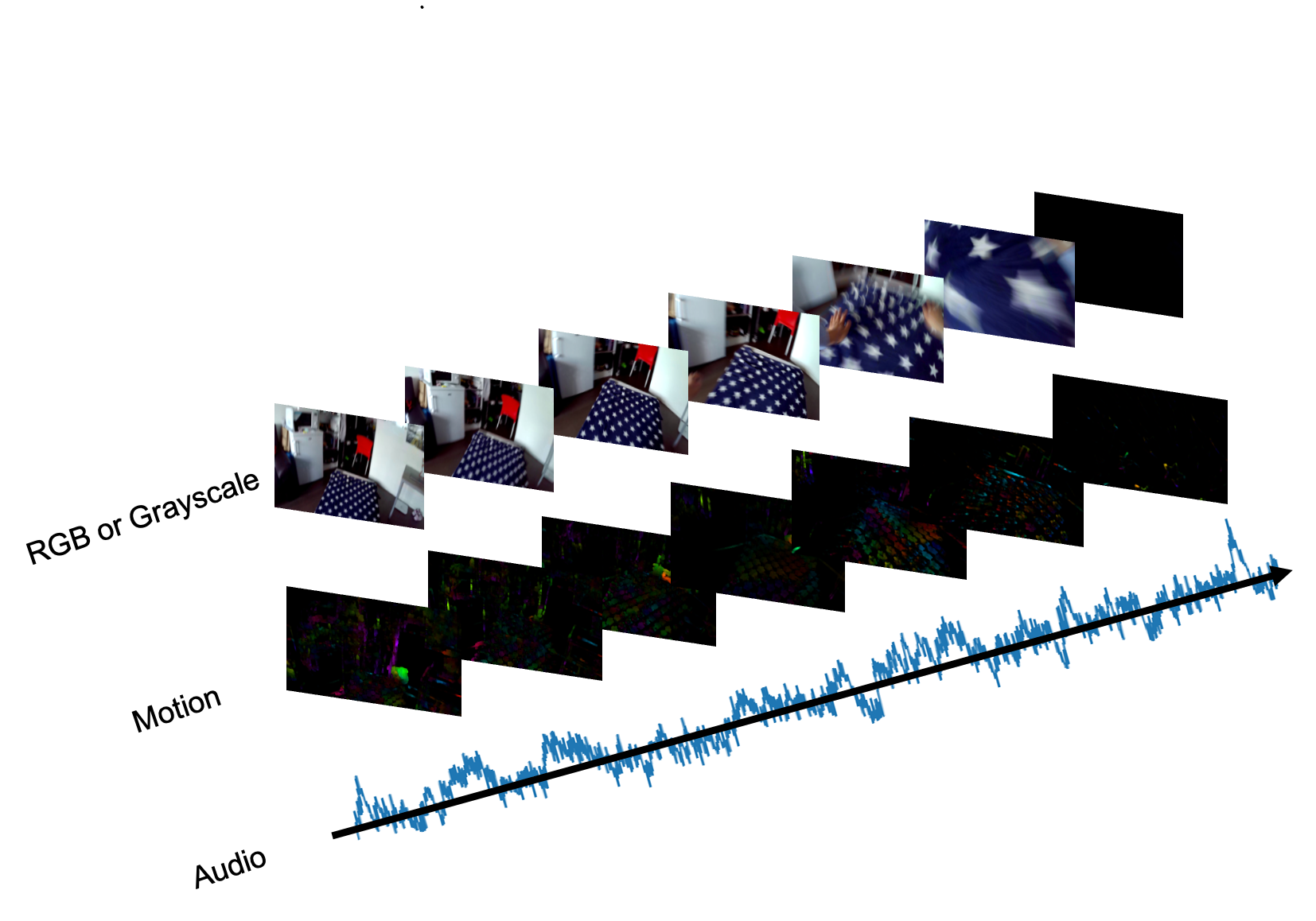}
    \caption{Three streams of spatial, motion of vision, and audio by video clips.}
    \label{fig:multimodel}
\end{figure}
In the domain of fall detection, research has explored various sensing modalities which can be categorized into four primary types: wearable sensors, fixed visual sensors, ambient sensors, and sensor fusion \cite{wang2020elderly}. Wearable sensors are a popular choice due to their portability and capability to capture data without spatial constraints. Furthermore, they offer the advantage of directly recording physiological changes associated with the human body. Fixed visual sensors, on the other hand, are valued for their simplified hardware, high-quality imaging, and reliable performance. Various forms of visual sensors have been investigated, including fixed RGB cameras and RGB-D depth cameras \cite{ma2014depth} and wearable cameras~\cite{wang2021fall, Xueyi2023}. The integration of wearable and visual sensors has given rise to wearable cameras as a promising technology for fall detection. 

Huang et al. \cite{huang2023egocentric} proposed an innovative approach for egocentric visual-audio object localization, addressing the challenges posed by egomotion and out-of-view auditory components. Their proposed method comprises a geometry-aware temporal aggregation module, a cascaded feature enhancement module, and a soft localization module. The EPIC-Fusion study by Kazakos et al. \cite{kazakos2019epic} introduces a novel architecture for multi-modal temporal binding, integrating RGB, Flow, and Audio modalities for the purpose of egocentric action recognition. The method achieved state-of-the-art results on the largest egocentric dataset, EPIC-Kitchens, and underscores the significance of auditory input in egocentric cameras for the identification of actions and interactive objects.

In a separate study, Xiao et al. \cite{xiao2020audiovisual} presented a novel architecture termed Audiovisual SlowFast Networks (AVSlowFast) designed for integrated audiovisual perception in video recognition. The AVSlowFast architecture fuses audio and visual features at multiple levels to form a unified representation, incorporating a novel DropPathway technique to mitigate training difficulties. The study establishes the effectiveness of the AVSlowFast architecture for both supervised and self-supervised learning of audiovisual features, offering promising prospects for advancements in video analysis applications.

Despite the extensive research in fall detection, previous studies have not offered a solution tailored to low-illumination conditions. In contrast, our study employs egocentric cameras and fuses RGB, infrared visual, and audio data to detect falls under varying illumination levels, addressing a gap that has remained unexplored in prior research.
The contributions of this paper is listed as follows:
\begin{itemize}
    \item We have assembled an extensive multimodal dataset comprising RGB and infrared videos, as well as audio recordings, for the purpose of investigating fall detection. This dataset also holds potential for application in research focused on fall related activity recognition using egocentric cameras.
    \item We developed a late decision model that could be detect falls successfully in both low and high illumination environment as the benchmark for this dataset.
\end{itemize}

We organize the rest of the paper as follows. We introduce how to design the data collection in Section 2. Then we depict the methods in Section 3, show the experimental results and conduct a discussion in Section 4. We draw some conclusions in the last section 5.

\begin{table}[ht!]
\centering
\caption{Quantity and type of video clips per participant, where C1 and C2 refers to camera 1 and camera 2, and 0 means that no videos were collected for the corresponding activities.}
\resizebox{\columnwidth}{!}{
\begin{tabular}{llcccccccc}
\toprule
\textbf{Data} & \textbf{ID} & \textbf{Camera/Time} & \textbf{All} & \textbf{Non-Falls} & \textbf{Falls} & \textbf{Indoor} & \textbf{Outdoor} & \textbf{Waist} & \textbf{Neck}\\
\midrule
P1 & S\_H & C1-RGB/daytime &1096 &328 & 768 & 554 & 542 & 548 & 548 \\

\midrule
P2 & S\_M & C1-RGB/daytime & 938 & 426 & 512 & 562 & 376 & 469 & 469\\

\midrule
P3 & S\_R & C1-RGB/daytime & 1630 & 680 & 950 & 812 & 818 & 815 & 815\\

\midrule
P4 & S\_W & C1-RGB/daytime & 1298 & 536 & 762 & 586 & 712 & 649 & 649\\

\midrule
P5 & S\_XL & C1-RGB/daytime & 896 & 444 & 452 & 374 & 522 & 448 & 448\\
\midrule

P6 & S\_Q & C1-RGB/daytime & 658 & 498 & 160 & 346 & 312 & 329 & 329\\
\midrule

P7 & S\_FI & C1-RGB/daytime & 208 & 136 & 72 & 116 & 92 & 104 & 104\\
\midrule

P8 & S\_HB & C1-RGB/daytime & 490 & 316 & 174 & 278 & 212 & 245 & 245\\
\midrule

P9 & S\_F & C1-RGB/daytime & 142 & 142 & 0 & 142 & 0 & 71 & 71\\
\midrule

P10 & S\_JF & C1-RGB/daytime & 148 & 148 & 0 & 148 & 0 & 74 & 74\\
\midrule

P11 & S\_L & C1-RGB/daytime & 380 & 217 & 163 & 248 & 132 & 190 & 190\\
\midrule

P12 & S\_D\_W & C1-RGB/night & 446 & 318 & 128 & 246 & 200 & 223 & 223\\
\midrule

P13 & S\_D\_WD & C1-RGB/night & 394 & 264 & 130 & 186 & 208 & 197 & 197\\
\midrule

P14 & S\_I\_R & C2-Infra/night & 500 & 500 & 0 & 196 & 304 & 250 & 250\\

\midrule
P15 & S\_I\_W & C2-Infra/night & 454 & 336 & 118 & 230 & 224 & 227 & 227\\
\midrule

P16 & S\_I\_ZJ & C2-Infra/night & 628 & 444 & 184 & 316 & 312 & 314 & 314\\
\midrule

P17 & S\_I\_CZ & C2-Infra/night & 642 & 478 & 164 & 322 & 320 & 321 & 321\\
\midrule

\textbf{All} & \textbf{All} & \textbf{All}& \textbf{10948} & \textbf{7177} & \textbf{3771} & \textbf{5628} & \textbf{5320} & \textbf{5474} & \textbf{5474} \\
\bottomrule
\end{tabular}
}
\label{number of clips}
\end{table}

\section{Dataset}
The majority of earlier data sets lacked certain crucial elements present in the actual world, such as various lighting, various subjects, various environments, and various camera locations. In this data collection, we take them into account. We will use cameras, subjects, environments, and guidelines for data simulation to explain specifics of data collecting. The dataset is the largest compared to other datasets of falls collected from egocentric cameras; 330 clips in \cite{casares2012automatic} and 237 in \cite{ozcan2013automatic}. The dataset will be publicly available\footnote{https://dataverse.nl/dataset.xhtml?persistentId=doi:10.34894/HO5GE3}. 

\textbf{Equipment:} Data was collected using two types of wearable cameras: the OnReal G1 and CAMMHD Bodycams. The OnReal G1 is a compact mini action camera measuring $420 \times 420 \times 200$mm, capable of capturing high-resolution videos up to $1080$P at 30 fps. On the other hand, the CAMMHD Bodycam is a larger body camera measuring $800 \times 500 \times 300$mm, equipped with infrared sensors for night vision. These cameras were strategically attached to various locations on the human body, such as the waist and neck, to gather comprehensive visual, motion and audio information from different environments as shown in~Fig.\ref{fig:multimodel}. For data collection, the chosen settings were 1080p video mode at 30 frames per second. Notably, the OnReal G1 frames include three distinct R, G, B channels, while CAMMHD Bodycam frames consist of three identical, repetitive grayscale channels. The resulting dataset serves as a valuable resource for this thesis, enabling a comprehensive analysis of events and activities.

\textbf{Subject:} In this data collection study, 14 volunteers participated, consisting of 12 males and 2 females, with 12 young healthy individuals and 2 elderly subjects. All participants provided their consent, acknowledging that their data would be used for research and potentially made public. The majority of subjects (11 out of 14) successfully completed data collection for four types of falls and nine types of non-falls, both indoors and outdoors. However, three subjects were unable to complete all data collection due to personal reasons. This research offers valuable insights into falls and non-falls behavior and highlights the commitment and dedication of the majority of participants to the study.

\textbf{Environment:} We aim to comprehensively address both indoor and outdoor environments by encompassing 14 distinct common outdoor settings and 15 diverse indoor spaces to gather data from all subjects. To enhance the variety of environmental conditions, participants are encouraged to alter their positions or directions after each activity. By incorporating this approach, we ensure a well-rounded and encompassing dataset, enabling us to draw more robust conclusions and insights for our research.

\textbf{Data collection:}
We explore data collection from two perspectives: vision and audio. For visual information, we adhere to the guidelines provided in \cite{abbate2010monitoring, yu2008approaches}. According to \cite{yu2008approaches}, falls and other fall-related activities typically last 1-3 seconds, and they proposed a comprehensive set of trials comprising 20 types of falls with various directions and interactions with different objects. In contrast, there are no specific guidelines for audio data, as previous studies have predominantly focused on visual information. Our audio dataset comprises three categories: subject audio, subject-object audio, and environment audio. To authentically recreate the sensory experience of falls, we expose participants to online videos depicting real-life incidents of individuals encountering genuine falls. These videos faithfully capture both the visual and auditory elements of these occurrences, allowing participants to immerse themselves in a true-to-life simulation of the event. Manual inspection of all clips helped identify common audio patterns. For falls, subject audio includes yelling, shouting, and moaning, subject-object audio captures sounds like hitting the ground or mattress, while environment audio encompasses ambient sounds such as traffic, wind, rain, animals, crowds (outdoor environment), and television, music, and talking (indoor and outdoor environments). Not every clip contains all these sounds, with some having none or a majority of them. Non-fall activities are categorized into two groups: those with strong subject-object audio (stumbling, walking, sitting-down, rising, lying), weak subject-object audio (bending, squatting down), and others without subject-object audio (sitting-static, standing). Notably, distinct sounds on the ground accompany stumbling and walking, while sounds of interacting with furniture accompany sitting down and rising. For bending and squatting, only friction sounds from cameras and clothes are noticeable. These findings provide valuable insights into audio patterns in various activities and can enhance future research in the field.

\section{Methods}
\textbf{Visual descriptors by handcrafted features.} 
In the context of this study, we employ three types of handcrafted feature descriptors, namely Histogram of Oriented Gradients (HOG), Local Binary Patterns (LBP), and Optical Flow. HOG, predominantly utilized in object detection tasks, quantifies the occurrences of varied gradient orientations within localized regions of a given image. The LBP descriptor \cite{ojala1994performance}, conversely, characterizes the neighborhood of image elements using binary codes. It captures diverse features including edges, lines, spots, and flat areas, by leveraging two complementary measures: local spatial patterns and grayscale contrast. Lastly, Optical Flow, our third handcrafted feature, measures the apparent motion between two consecutive video frames at each position, thereby providing a detailed analysis of the temporal changes in the video frames.

\textbf{Truncating and aligning video descriptors.} We computed the similarity for three types of handcrafted feature descriptors using two vectors of size $k=1152$, extracted from consecutive frames, resulting in a vector of $n-1$ elements for each video, each representing the cosine similarity as shown in Eq~\ref{equ:cos}. In our system, we process videos within a user-defined time window of at least 8 seconds, adjustable for specific needs, and from our dataset of 8 to 40-second-long video clips, we truncate each video's descriptors to a uniform length of 238 elements, centered around the maximum value within the time window.

\begin{equation}
\label{equ:cos}
  \text { cosine similarity }=\frac{\mathbf{A} \cdot \mathbf{B}}{\|\mathbf{A}\|\|\mathbf{B}\|}=\frac{\sum_{i=1}^{k} A_{i} B_{i}}{\sqrt{\sum_{i=1}^{k} A_{i}^{2}} \sqrt{\sum_{i=1}^{k} B_{i}^{2}}},
\end{equation}

\textbf{Visual descriptors by deep features.} In this study, we harness the power of deep features extracted from video frames using the pre-trained ResNet-50 model \cite{he2016deep}, renowned for its robustness against network degradation and vanishing gradients due to its deep residual network architecture and skip connections. We resize each video frame to $224 \times 224$ pixels, extract a 2048-element vector from the network's last fully connected layer as a global image descriptor, and, for computational efficiency, represent each clip with a 20,480-element vector by concatenating the feature descriptors from ten equally spaced frames.

\textbf{Audio features.}  
The Mel-frequency Cepstral Coefficients (MFCC) feature extractor in the Librosa library is a widely used method for audio signal processing, transforming audio signals into compact representations that consider the characteristics of the human auditory system. This process involves converting the audio signal to the frequency domain, applying a Mel filter bank that mimics human ear perception, taking logarithms of the filter bank energies, and using discrete cosine transform to decorrelate the energies and reduce feature vector dimensionality, thus capturing perceptual content and yielding robust features against noise, pitch, and amplitude variations.
\begin{equation}
\label{equ:e3}
\small
\begin{split}
DecisionFusion([p^H_1,\dots,p^H_m], [p^L_1,\dots,p^L_m],[p^O_1,\dots,p^O_m],\\ [p^D_1,\dots,p^D_m]) = [P_1,\dots,P_m],
\end{split}
\end{equation}
\textbf{Decision fusion and classification.}

In this study, we employ a decision fusion technique to integrate the classification results of four independent models, each trained on distinct video descriptors encompassing handcrafted features (HOG, LBP, optical flow), deep features (ResNet50), and audio features. The output vectors from Models 1 through 5 are amalgamated according to Eq~\ref{equ:e3}. We applied RF, SVM, SVM, MLP for handcraft visual features, deep visual features, audio features, and late decision fusion separately in this work based on the features of each classifiers and evaluation of previous work~\cite{wang2021fall,Xueyi2023, wang2023fall}. 

\section{Experiments}

\subsection{Implementation details}

\textbf{Evaluation metrics.} 
We evaluated the generalization ability of our machine learning models using both internal and external cross-validation techniques. 

\noindent \textbf{Experimental setup.} 
Internal cross-validation, implemented by dividing the entire dataset into $k$ subsets or folds, was used to assess model stability and robustness within the dataset by training and validating the model $k$ times, rotating the validation set each time, and calculating average performance metrics. In contrast, \textit{external cross-validation} employed a leave-one-subject-out approach, simulating real-world situations where the model encounters unseen data, aiding in model selection, hyper-parameter tuning, and identifying over-fitting issues while also assessing predictive accuracy and adaptability to new instances, thus verifying the model's overall effectiveness. Data from participants of P9, P10, and P14 were excluded from the analysis due to incomplete data collection.

\subsection{Results and discussion }

\textbf{Internal cross-validation}.
The results of our internal cross-validation are presented in Table~\ref{table:internal}. Our approach, fusing audio and vision, yielded the highest performance for binary classification with an accuracy of 0.978, and for 12-class classification with an accuracy of 0.850. Notably, fusing four types of individual visual feature extractors led to higher accuracy than using each visual feature descriptor independently. The performance of the audio feature extractor in binary classification was comparable to that of the fused visual feature extractors, though it slightly underperformed in the 12-class classification scenario.

\begin{table}[ht]
\footnotesize
    \centering
    \caption{The results for six baseline models using internal evaluation for both the 2-class (fall detection) and 12-class (daily activity recognition) problems. B1 refers to five kinds of individual features (HOG, LBP, optical flow, resnet50, audio), B2 refers to fusion of visual features, and the proposed fusion approach ``Ours" includes all features.}
    \scalebox{0.94}{
    \begin{tabular}{c c c p{2cm} p{1.8cm}} 
    \toprule 
    \textbf{Fusion} & \textbf{Handcrafted} & \textbf{Deep} &  \multicolumn{2}{c}{\textbf{Accuracy}} \\ 
    \textbf{features} & \textbf{features} & \textbf{model} & \textbf{2 classes} & \textbf{12 classes}\\
    \midrule

    B1 & HOG & $\times$ & 0.785 ($\pm$ 0.01) & 0.473 ($\pm$ 0.03)\\ 
    B1 & LBP & $\times$ & 0.855 ($\pm$ 0.02) & 0.555 ($\pm$ 0.00)\\ 
    B1 & Optical flow & $\times$ & 0.843 ($\pm$ 0.01) & 0.536 ($\pm$ 0.01)\\ 
    B1 & Resnet50 & \checkmark &  0.955 ($\pm$ 0.01) & 0.594 ($\pm$ 0.01)\\ 
    B1 & Audio & $\times$ &  0.952 ($\pm$ 0.01) & 0.730 ($\pm$ 0.01)\\ 
    B2 & Vision & \checkmark &  0.952 ($\pm$ 0.01) & 0.742 ($\pm$ 0.01)\\ 
    \midrule
    \textbf{Ours} & All &  \checkmark & \textbf{0.978 ($\pm$ 0.01)}& \textbf{0.850 ($\pm$ 0.01)}\\      
  \bottomrule
    \end{tabular}
}
    \label{table:internal}
\end{table}

\textbf{External cross-validation}. 
In this study, the generalization of our model was evaluated using external cross-validation, where data from one subject was designated as the test set while data from all other subjects served as the training set. The late decision fusion model, which integrates vision and audio, demonstrated the better performance, consistent with the findings in internal cross-validation. However, a decrease in performance relative to the internal cross-validation was observed. Despite this, the model achieved respectable accuracy scores of 0.875 for binary classification and 0.520 for 12-class classification in external cross-validation. Moreover, as some data were collected in a dark environment, results were reported separately for low and high illumination using RGB and infrared cameras. Specifically, for RGB in high illumination (daytime), an accuracy of 0.924 was achieved for fusion of all features. For low illumination, the RGB cameras achieved 0.746 accuracy, while infrared cameras achieved 0.913 accuracy for fusion of vision and audio. For vision fusion, the accuracies were 0.883 for RGB in high illumination, 0.751 for RGB in low illumination, and 0.805 for infrared in low illumination. For audio, the accuracies were 0.902 for RGB in high illumination, 0.891 for RGB in low illumination, and 0.827 for infrared in low illumination. While high accuracy was maintained across both low and high illumination for audio information, the results indicate that the vision feature extractor exhibited reduced performance in low illumination. 
\begin{table}[t]
\footnotesize
    \centering
    \caption{The results for six baseline models using external evaluation for both the 2-class (fall detection) and 12-class (daily activity recognition) problems. B1 refers to five kinds of individual features (HOG, LBP, optical flow, resnet50, audio), B2 refers to fusion of visual fetures, and the proposed fusion approach ``Ours" include all features.}
    \scalebox{0.94}{
    \begin{tabular}{c c c p{2cm} p{1.8cm}} 
    \toprule 
    \textbf{Fusion} & \textbf{Handcrafted} & \textbf{Deep} &  \multicolumn{2}{c}{\textbf{Accuracy}} \\ 
    \textbf{features} & \textbf{features} & \textbf{model} & \textbf{2 classes} & \textbf{12 classes}\\
    \midrule

    B1 & HOG & $\times$ & 0.746 ($\pm$ 0.05) & 0.389 ($\pm$ 0.21)\\ 
    B1 & LBP & $\times$ & 0.804 ($\pm$ 0.13) & 0.427 ($\pm$ 0.35)\\ 
    B1 & Optical flow & $\times$ & 0.801 ($\pm$ 0.07) & 0.437 ($\pm$ 0.33)\\ 
    B1 & Resnet50 & \checkmark &  0.789 ($\pm$ 0.14) & 0.343 ($\pm$ 0.26)\\ 

    B1 & Audio & $\times$ &  0.886 ($\pm$ 0.10) & 0.395 ($\pm$ 0.15)\\ 
    B2 & Vision & \checkmark &  0.845 ($\pm$ 0.12) & 0.496 ($\pm$ 0.35)\\ 
    \midrule
    \textbf{Ours} & All &  \checkmark & \textbf{0.875 ($\pm$ 0.15)}& \textbf{0.520 ($\pm$ 0.33)}\\      
  \bottomrule
    \end{tabular}
    }
    \label{table:external}
\end{table}

\section{Conclusion}
This work presents a new dataset and benchmark in the field of fall detection. We proposed a multimodal learning approach to detect falls in first-person view videos collected by egocentric cameras. The descriptors we explore are images, motion, and audio. The obtained results indicate that the proposed late decision fusion model, which combines visual and auditory data, has the capability for detecting falls under a wide range of lighting conditions, encompassing both high and low illumination scenarios during both daytime and nighttime.


\bibliographystyle{IEEEbib}
\bibliography{strings,refs}

\end{document}